\documentclass[10pt,twocolumn,letterpaper]{article}

\usepackage{subcaption}
\usepackage{float}
\usepackage{enumitem}
\setlist[enumerate]{itemsep=0mm}
\usepackage{xspace}
\usepackage[export]{adjustbox}

\usepackage{bm}
\usepackage{mathtools}

\usepackage{multirow}

\usepackage{units}
\usepackage{color}

\definecolor{darkgreen}{RGB}{0, 153, 76}
\definecolor{darkyellow}{RGB}{255, 128, 0}

\usepackage{iccv}
\usepackage{times}
\usepackage{epsfig}
\usepackage{graphicx}
\usepackage{amsmath}
\usepackage{amssymb}

\usepackage{booktabs}

\DeclareMathAlphabet{\altmathcal}{OMS}{cmsy}{m}{n}
\DeclareMathAlphabet{\mathbfit}{OT1}{ptm}{bx}{it}

\newcommand {\first}[1]{{\textbf{#1}}}
\newcommand {\second}[1]{{\underline{#1}}}

\newcommand{\mpage}[2]
{
\begin{minipage}{#1\linewidth}\centering
#2
\end{minipage}
}

\newbox\jsavebox%

\makeatletter
\newcommand{\providelength}[1]{%
  \@ifundefined{\expandafter\@gobble\string#1}
   {
    \typeout{\string\providelength: making new length \string#1}%
    \newlength{#1}%
   }
   {
    \sdaau@checkforlength{#1}%
   }%
}

\newcommand{\sdaau@checkforlength}[1]{%
  \edef\sdaau@temp{\expandafter\sdaau@getfive\meaning#1TTTTT$}%
  \ifx\sdaau@temp\sdaau@skipstring
    \typeout{\string\providelength: \string#1 already a length}%
  \else
    \@latex@error
      {\string#1 illegal in \string\providelength}
      {\string#1 is defined, but not with \string\newlength}%
  \fi
}
\def\sdaau@getfive#1#2#3#4#5#6${#1#2#3#4#5}
\edef\sdaau@skipstring{\string\skip}
\makeatother

\newcommand{\fgnet}{f_{\text{fg}}}
\newcommand{\bgnet}{f_{\text{bg}}}
\newcommand{\fgmask}{M^\text{fg}_t}
\newcommand{\bgflow}{F_t^\text{bg}}
\newcommand{\maskedflow}{F^{\text{m}}_t}

\newcommand{\Loss}{\altmathcal{L}}
\newcommand{\Lrecons}{\Loss_{\text{recons}}}
\newcommand{\Lalphareg}{\Loss_{\alpha\text{-reg}}}
\newcommand{\Lalphawarp}{\Loss_{\alpha\text{-warp}}}
\newcommand{\Lmask}{\Loss_{\text{mask}}}
\newcommand{\Lflowrecons}{\Loss_{\text{flow}}}
\newcommand{\Ldepth}{\Loss_{\text{depth}}}
\newcommand{\Ldistort}{\Loss_{\text{distort}}}
\newcommand{\Lbgreg}{\Loss_{\text{bg-reg}}}

\newcommand{\dsformat}[1]{\texttt{#1}}
\newcommand{\dskubric}{\dsformat{Kubrics}}
\newcommand{\dsmovies}{\dsformat{Movies}}
\newcommand{\dsdavis}{\dsformat{DAVIS}}
\newcommand{\dswild}{\dsformat{Wild}}
\newcommand{\dtnerf}{D$^2$NeRF}

\newcommand{\updated}[1]{#1}

\usepackage[pagebackref=true,breaklinks=true,letterpaper=true,colorlinks,bookmarks=false]{hyperref}

\iccvfinalcopy 

\def\iccvPaperID{11144} 

\ificcvfinal\fi

\begin{document}

\title{OmnimatteRF: Robust Omnimatte with 3D Background Modeling}

\author{Geng Lin$^1$ \qquad
Chen Gao$^2$ \qquad
Jia-Bin Huang$^{1,2}$ \vspace{0.1em} \\
Changil Kim$^2$ \qquad
Yipeng Wang$^2$ \qquad
Matthias Zwicker$^1$ \qquad
Ayush Saraf$^2$
\vspace{0.2em}
\\
$^1$University of Maryland, College Park \qquad $^2$Meta
\vspace{0.1em}
\\
{\small\url{https://omnimatte-rf.github.io}}
}

\maketitle
\ificcvfinal\thispagestyle{empty}\fi

\begin{abstract}
Video matting has broad applications, from adding interesting effects to casually captured movies to assisting video production professionals.
Matting with associated effects such as shadows and reflections has also attracted increasing research activity, and methods like Omnimatte have been proposed to separate dynamic foreground objects of interest into their own layers.
However, prior works represent video backgrounds as 2D image layers, limiting their capacity to express more complicated scenes, thus hindering application to real-world videos.
In this paper, we propose a novel video matting method, OmnimatteRF, that combines dynamic 2D foreground layers and a 3D background model.
The 2D layers preserve the details of the subjects, while the 3D background robustly reconstructs scenes in real-world videos.
Extensive experiments demonstrate that our method reconstructs scenes with better quality on various videos.
\end{abstract}


\section{Introduction}
\label{sec:intro}

\newcommand{\omparallaxscene}{wild/car}
\begin{figure}[t]
    \centering
    \captionsetup[subfigure]{labelformat=empty}

    \begin{subfigure}{.49\linewidth}
    \centering
    \captionsetup{width=0.98\linewidth}
    \includegraphics[width=0.98\linewidth]{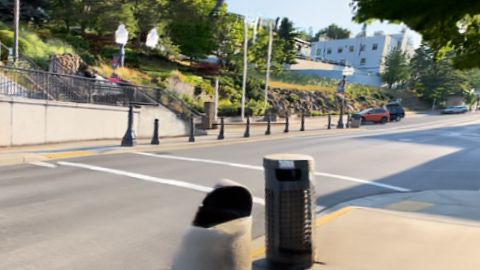}
    \caption{{\footnotesize (a) Omnimatte BG}}
    \end{subfigure}
    \begin{subfigure}{.49\linewidth}
    \centering
    \captionsetup{width=0.98\linewidth}
    \includegraphics[width=0.98\linewidth]{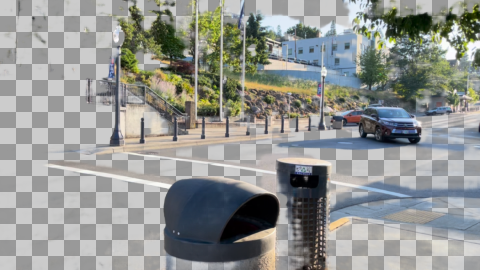}
    \caption{{\footnotesize (b) Omnimatte FG}}
    \end{subfigure}
    \begin{subfigure}{.49\linewidth}
    \centering
    \captionsetup{width=0.98\linewidth}
    \includegraphics[width=0.98\linewidth]{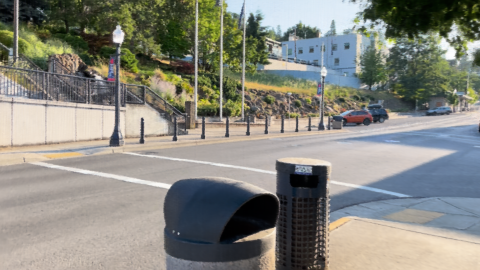}
    \caption{{\footnotesize (c) Our BG}}
    \end{subfigure}
    \begin{subfigure}{.49\linewidth}
    \centering
    \captionsetup{width=0.98\linewidth}
    \includegraphics[width=0.98\linewidth]{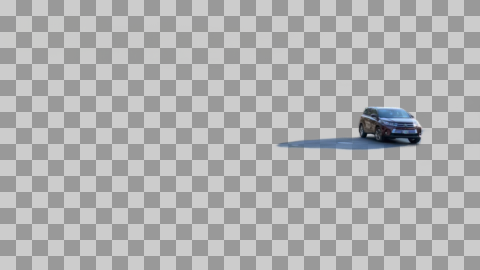}
    \caption{{\footnotesize (d) Our FG}}
    \end{subfigure}

\centering
\caption{
\textbf{Video with parallax effects.} Limited by their 2D image representation (a), previous works such as Omnimatte fail to handle videos with parallax effects in the background. Their foreground layer (b) has to capture (dis)occlusion effects to minimize the reconstruction loss. In contrast, our method employs a 3D background (c), enabling us to obtain clean foreground layers (d).
}
\label{fig:om_results}
\end{figure}

Video matting is the problem of separating a video into multiple layers with associated alpha mattes such that the layers are composited back to the original video.
It has a wide variety of applications in video editing as it allows for substituting layers or processing them individually before compositing back, and thus has been studied well over decades. In typical applications like rotoscoping in video production and background blurring in online meetings, the goal is to obtain the masks containing only the object of interest. In many cases, however, it is often preferred to be able to create video mattes that include not only the object of interest but also its associated effects, like shadow and reflections. This could reduce the often-required, additional manual segmentation of secondary effects and help increase realism in the resulting edited video.
Being able to factor out the related effects of foreground objects also helps reconstruct a clean background, which is preferred in applications like object removal. Despite these benefits, this problem is much more ill-posed and has been much less explored than the conventional matting problem.

The most promising attempt to tackle this problem is Omnimatte~\cite{omnimatte}. \textit{Omnimattes} are RGBA layers that capture dynamic foreground objects and their associated effects. Given a video and one or more coarse mask videos, each corresponding to a foreground object of interest, the method reconstructs an \textit{omnimatte} for each object, in addition to a static background that is free from all of the objects of interest \emph{and} their associated effects.
While Omnimatte~\cite{omnimatte} works well for many videos, it is limited by its use of homography to model backgrounds, which requires the background be planar or the video contains only rotational motion. This is not the case as long as there exists parallax caused by camera motions and objects occlude each other. This limitation hinders its application in many real-world videos, as shown in Fig. \ref{fig:om_results}.

\dtnerf{}~\cite{d2nerf} attempts to address this issue using two radiance fields, which model the dynamic and static part of the scene. The method works entirely in 3D and can handle complicated scenes with significant camera motion. It is also self-supervised in the sense that no mask input is necessary. However, it separates all \emph{moving} objects from a static background and it is not clear how to incorporate 2D guidance defined on video such as rough masks. Further, it cannot independently model multiple foreground objects. A simple solution of modeling each foreground object with a separate radiance field could lead to excessive training time, yet it is not clear how motions could be separated meaningfully in each radiance field.

We propose a method that has the benefit of both by combining 2D foreground layers with a 3D background model. The lightweight 2D foreground layers can represent multiple object layers, including complicated objects, motions, and effects that may be challenging to be modeled in 3D. At the same time, modeling background in 3D enables handling background of complex geometry and non-rotational camera motions, allowing for processing a broader set of videos than 2D methods.
We call this method \emph{OmnimatteRF} and show in experiments that it works robustly on various videos without per-video parameter tuning. To quantitatively evaluate the background separation of a 3D scene, \dtnerf{} released a dataset of 5 videos rendered with Kubrics, which are simple indoor scenes with few pieces of furniture and some moving objects that cast solid shadows.

We also render five videos from open-source Blender movies \cite{blenderStudio} with sophisticated motions and lighting conditions for more realistic and challenging settings. 
Our method outperforms prior works in both datasets, and we release the videos to facilitate future research.

In summary, our contributions include the following:

\begin{enumerate}
	\item We propose a novel method to make Omnimatte \cite{omnimatte} more robust by better modeling the static background in 3D using radiance fields \cite{mildenhall2020nerf}.
	\item Utilizing the \textit{omnimatte} masks, we propose a simple yet effective re-training step to obtain a clean static 3D reconstruction from videos with moving subjects.
	\item We release a new dataset of 5 challenging video sequences rendered from open-source blender movies \cite{blenderStudio} with ground truths to better facilitate the development and evaluation of the video matting with associated effects (aka \textit{omnimatting} \cite{omnimatte}) problem.
\end{enumerate}

\section{Related Work}
\label{sec:related}

\paragraph{Video Matting.}
There is a long line of work exploring video matting due to its importance in video editing. Green screening and rotoscoping are critical first steps in any visual effects pipeline. The matting problem aims to extract the foreground subjects into their own RGBA layers and separate them from the background RGB layer, which is a highly under-constrained problem. Many approaches have utilized motion and depth cues in addition to integrating user interactions~\cite{brostow1999motion,bai2009video,wang2005interactive,li2005video,chuang2002video}. Background Video Matting \cite{BGMv2} specifically addresses real-time video matting of people and preserving strand-level hair details.

\paragraph{Matting with Associated Effects.}
Video matting is often insufficient, as foreground subjects might have associated effects like shadows or reflections that need to be extracted into the foreground RGBA layers. This problem has not been explored as extensively and, in practice, is often dealt with manually using advanced interactive rotoscoping tools \cite{Rotopp2016}. Omnimatte \cite{omnimatte} was the first to propose a generic framework capable of learning any associated effect. Previous works often specifically addressed associated effects like shadows \cite{Wang_2020_CVPR,Wang_2021_CVPR}. The ability to obtain matte layers with associated effects has many exciting applications, such as re-timing motions of different people \cite{lu2020retiming}, consistent background editing \cite{layeredNeuralAtlas,lee2023shape}, background subtraction, green screening, and many other video effects \cite{omnimatte}. 
Recently, FactorMatte \cite{factormatte} has been proposed to improve the quality with data augmentation and conditional priors. 
These works have in common that they take predefined masks that hint at the foreground objects and decompose each video into several layers, with one object in each layer with its associated effects. Then, there is a background layer, a 2D static image, or a deformable atlas shared by all the frames. The background is warped and cropped via a homography to render each frame. While the foreground layers have shown great potential in capturing dynamics, their single image background limits the application of these methods to videos with planar environments without parallax effects caused by camera motion.

\paragraph{Radiance Fields.}
Radiance fields (RF) emerged as 3D representations capable of capturing geometric details and photorealistic appearances~\cite{mildenhall2020nerf}. Radiance fields model the 3D scene as a continuous function that maps the position and the viewing direction of any point in world space to its color and opacity. Novel views can be synthesized via volume rendering along rays cast. This continuous function is learned by optimizing with a reconstruction loss on the rendered images. This view-dependent volumetric representation can model various challenging scenes that previous surface-based methods struggled to handle: e.g., shiny surfaces like metals or fuzzy surfaces like hair or fur. Since then, it has been extended along multiple axes: better appearance modeling (e.g., reflection and refraction \cite{verbin2022refnerf,bemana2022eikonal,attal2022learning,attal2023hyperreel}, faster optimization \cite{tensorf,SunSC22dvgo,mueller2022instant} and modeling dynamic scenes \cite{xian2021spacetime,li2021neural,gao2021dynamic,liu2023robust}.
Since the MLP-based implicit RF representations are slow to train, we use voxel-based explicit radiance field representations \cite{tensorf} \cite{SunSC22dvgo}. 
Specifically, we use the factorized voxel grid representation from \cite{tensorf}.

\paragraph{Self-Supervised Video Dynamics Factoring.}
Another related work is video dynamics factoring without needing a predefined mask. One recent work is deformable sprites \cite{deformableSprites} that rely only on motion cues. Similar to other video matting works, it has a 2D foreground and background layers and the same limitations as Omnimatte. For modeling in 3D, \dtnerf \cite{d2nerf} proposes to decouple the scene with two radiance fields, one for the dynamic content and the other for the statics. \dtnerf \cite{d2nerf} handles a special case of matting with only one foreground object, and, compared to the other methods, it is not limited to planar backgrounds. However, the self-supervised method relies on the heuristics that require per-video hyper-parameter tuning and does not robustly generalize to new videos. The quality of the foreground reconstruction can also be limited for objects that have large nonrigid motions.

We, therefore, propose a method for video matting with associated effects that has the advantages of supervised 2D mattes that support multiple individual objects with great details, as well as 3D background decoupling that works with non-planar videos.
\section{Method}
\label{sec:method}

\providelength\width
\setlength\width{2cm}

\begin{figure*}[t]
    \centering
    \includegraphics[width=\linewidth]{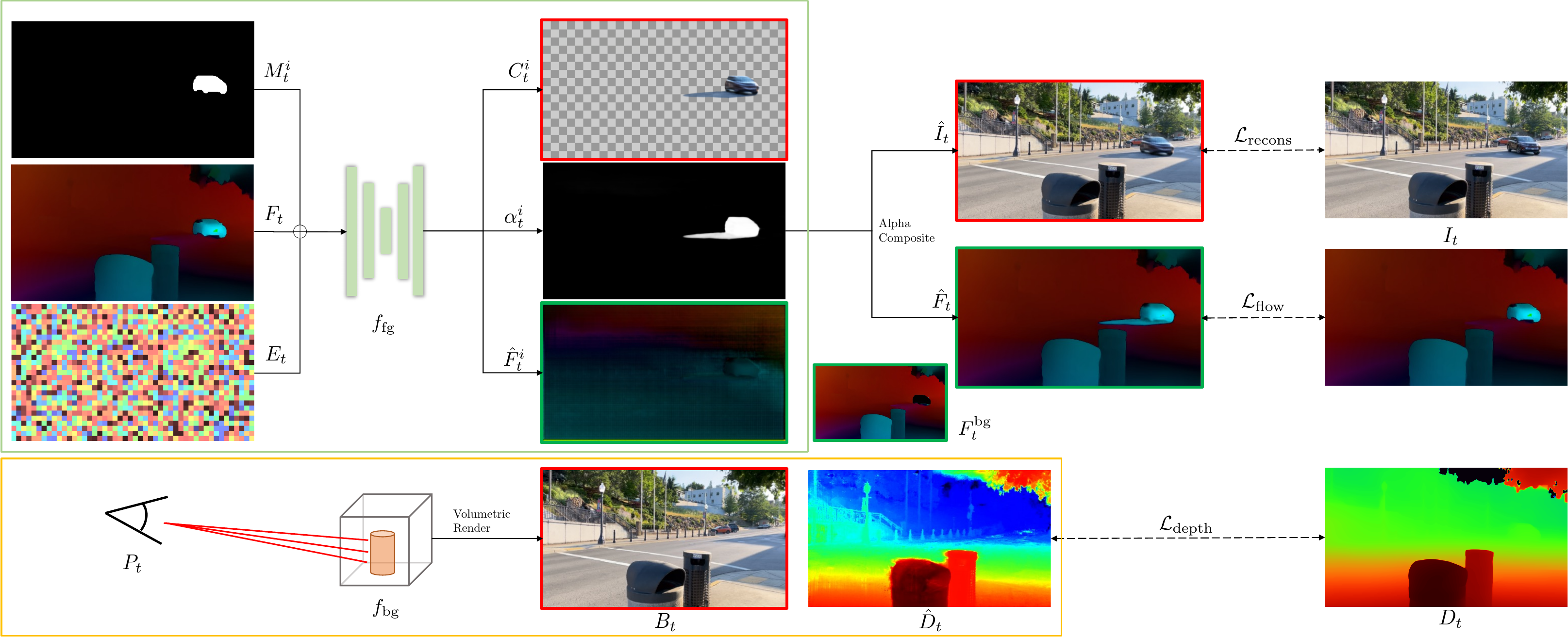}
    \caption{
    \textbf{Method overview.} We propose a video matting method, named OmnimatteRF, which combines 2D foreground layers with a 3D background layer. The foreground branch ($\fgnet$, in {\color{darkgreen}green} box) predicts an RGBA layer ($C_t^i, \alpha_t^i$) for each object, and an auxiliary flow output ($\hat{F}_t^i$). The background branch ($\bgnet$, in {\color{darkyellow}yellow} box) produces a background layer with depths ($B_t, \hat{D}_t$). \textbf{Optimization.} During training, predicted colors ($\hat{I}_t$) and flow ($\hat{F}_t$) are alpha-composited, whose inputs have {\color{red}red} and {\color{darkgreen}green} borders respectively. The right most column illustrates the data terms in the loss function, and we omit the regularization terms in this illustration.
    }
    \label{fig:overview}
\end{figure*}

The concept of \textit{omnimattes} is proposed by Lu et al. \cite{omnimatte}, extending RGBA video mattes to capture associated effects of the objects of interest like shadows and reflections. To avoid any confusion, in the following text, we refer to their work as capital Omnimatte, and the resulting RGBA layers as italic \textit{omnimatte}.
In the matting setup, the user prepares a video of $T$ frames $\{I_t\}_{t=1}^{T}$, and $N$ \textit{ordered} mask layers $\{M^i_t\}_{i=1}^{N}$, each containing a coarse mask video of an object of interest. The video's camera parameters are also precomputed as $\{P_t\}$. 

The goal is to predict RGBA foreground layers $C_t^i$ and $\alpha_t^i$ that contain the objects together with their associated effects, and a background layer $B_{t}$ which is clean and free from the effects cast by the foreground objects. An input frame $I_t$ should be reconstructed by alpha compositing the foreground layers above the background.

In Omnimatte, the background is represented by a static 2D image and a homography transform $P_t$. To compose a frame, part of the static background is extracted according to the estimated homography $P_t$. The key idea of our work is to represent the static background in 3D using a radiance field, while keeping the foreground in 2D to better capture the dynamics of objects. We employ an explicit factorized voxel-based radiance field~\cite{tensorf} to model the background. In this case, $P_t$ represents a camera pose, and a background frame is rendered with volume rendering. Note that the foreground layers are still 2D videos. We refer to this combination as the OmnimatteRF model.

\subsection{The OmnimatteRF Model}

An outline of our model is depicted in Figure \ref{fig:overview}. The model has two independent branches: foreground and background. For any given frame, the foreground branch predicts an RGBA image (\textit{omnimatte}) for each object, and the background branch renders a single RGB image.

\textbf{Preprocessing}. Following similar works, we use an off-the-shelf model RAFT \cite{deng2020RAFT} to predict optical flow between neighboring frames. The flow is used as an auxiliary input and ground truth for supervision, denoted by $\{F_t\}$. We also use an off-the-shelf depth estimator MiDaS \cite{Ranftl2022} to predict monocular depth maps $\{D_t\}$ for each frame and use them as ground truth for the monocular depth loss.

\textbf{Background}. The background branch consists of a static neural radiance field, $\bgnet$, encoding the 3D representation of the scene. To render a pixel in a frame $I_t$, a ray is traced according to the estimated camera pose $P_t$, and the final RGB color is produced via volumetric rendering. The result of rendering the entire frame is $(B_t, \hat{D}_t)=\bgnet(P_t)$, where $B_t$ is an RGB image and $\hat{D}_t$ is a depth map.

\textbf{Foreground}. The foreground branch is a UNet-style convolutional neural network, $\fgnet$, similar to that of Omnimatte. The input of the network is a concatenation of three maps:

\begin{enumerate}
	\item The coarse mask $M_t^i$. The mask is provided by the user, outlining the object of interest. Mask values are ones if the pixels are inside the object.
	\item The optical flow $F_t$. It provides the network with motion hints. Note that the network also predicts an optical flow as an auxiliary task (detailed in Sec. \ref{sec:fglosses}).
	\item The feature map $E_t$. Each pixel $(x,y)$ in the feature map is the positional encoding of the 3-tuple $(x,y,t)$.
\end{enumerate}

Multiple foreground layers are processed individually. For the $i$-th layer, the network predicts the \textit{omnimatte} layer $(C_t^i,\alpha_t^i)$ and the flow $\hat{F}_t^i$.

\textbf{Detail Transfer}. For a tradeoff between image quality and training time, the foreground network typically produces a color layer with missing details when the alpha layers have captured sufficient associated effects. To boost the output quality, Omnimatte transfers details from input frames. We include the same process in our pipeline. Note that this is a post-processing step to produce final results, and does not apply to model optimization.

\subsection{Optimizing the Model}

We optimize an OmnimatteRF model for every video since both branches of our model are video-specific. To supervise learning, we employ an image reconstruction loss and several regularization losses.

\subsubsection{Reconstruction Loss}

We compute the reconstruction loss with the composed image $\hat{I}_t$ by alpha composition of foreground and background layers:

\begin{align}
	\hat{I}_t = \sum_{i=1}^{N} \left(\prod_{j=1}^{i-1}(1-\alpha_t^j) \alpha_t^i C_t^i \right) + \prod_{i=1}^{N}(1-\alpha_t^i) B_t
	\label{eq:compose}
\end{align}

And the reconstruction loss is the mean-squared-error between the predicted and input frame,

\begin{align}
	\Lrecons=||\hat{I}_t - I_t||^2
\end{align}

The reconstruction loss supervises both branches of our pipeline simultaneously. Limited by the computational cost of volumetric rendering, the background layer is rendered only at sparse random locations at each step, where $\Lrecons$ is computed for the composed pixel values.

\subsubsection{Foreground Losses}
\label{sec:fglosses}

We follow Omnimatte and include the alpha regularization loss $\Lalphareg$, alpha warp loss $\Lalphawarp$, and flow reconstruction loss $\Lflowrecons$. We also bootstrap the initial alpha prediction to match the input mask with the mask loss $\Lmask$, which is gradually decayed and disabled once its value drops below the threshold.

While most regularization terms in Omnimatte can be applied directly to our pipeline, the flow reconstruction loss is an exception. The formulation of the loss remains identical: given the per-layer flow prediction $\hat{F}_t^i$ and a background layer flows $\bgflow$, the complete flow $\hat{F}_t$ is composed via alpha composition (Eq. \ref{eq:compose}). Then, the loss is defined as:

\begin{align}
	\Lflowrecons=||(\hat{F}_t - F_t)\otimes \fgmask ||^2
\end{align}

Here, $\fgmask$ is the union of all foreground masks ($\{M_t^i\}$) for the frame $I_t$, and the loss is only evaluated at the location of \textit{input} coarse masks. The authors of Omnimatte have shown the effectiveness of this loss in their case, and we also demonstrate its importance in an ablation study.

However, it remains unclear how $\bgflow$ can be obtained. In Omnimatte, the background flow can be derived from image homography, which serves both as an input to the network and a background for composition. On the other hand, since our 3D background has only known camera poses but not depths, we cannot obtain background flows directly. Instead, we use the ground truth flow $F_t$ as network input to provide motion cues and a masked version of $F_t$ as background flow for composition. The masked flow is $\maskedflow=F_t \otimes (1 - \fgmask)$, which is the ground truth optical flow with the regions marked in the coarse masks set to zeros. $\otimes$ denotes elementwise multiplication. We find it crucial to use $\maskedflow$ rather than $F_t$ for composition, as the latter case encourages the network to produce empty layers with $\alpha^i_t$ equal to zero everywhere.

\subsubsection{Background Losses}

Apart from the reconstruction loss, the background network is supervised by the total variation regularization loss, $\Lbgreg$, as in TensoRF \cite{tensorf}. In addition, monocular depth supervision is used to improve scene reconstruction when the camera motions consist of rotation only:

\begin{align}
	\Ldepth = \text{metric}(D_t, \hat{D}_t),
\end{align}
where $\hat{D}_t$ is the estimated depth from volume rendering \cite{mildenhall2020nerf}, and the metric function is the scale-invariant loss from MiDaS \cite{Ranftl2022}. Also, we empirically find that $\Ldepth$ can introduce floaters, and employ the distortion loss $\Ldistort$ proposed in Mip-NeRF 360 \cite{mipnerf360} to reduce artifacts in the background.

\subsubsection{Summary} The combined loss for joint optimization is:

\begin{align}
\begin{split}
	\Loss = & \Lrecons +
\underbrace{\Lalphareg + \Lalphawarp + \Lflowrecons + \Lmask}_\text{Foreground} + \\
	& \underbrace{\Lbgreg + \Ldepth + \Ldistort}_\text{Background}
\end{split}
\end{align}

At every optimization step, $\Lrecons$ and background losses are evaluated at sparse random locations. Foreground losses are computed for the full image.

\subsection{Clean Background via Masked Retraining}

\providelength\width
\setlength\width{2cm}

\begin{figure}
\centering

\begin{tabular}{ccc}
\includegraphics[width=\width]{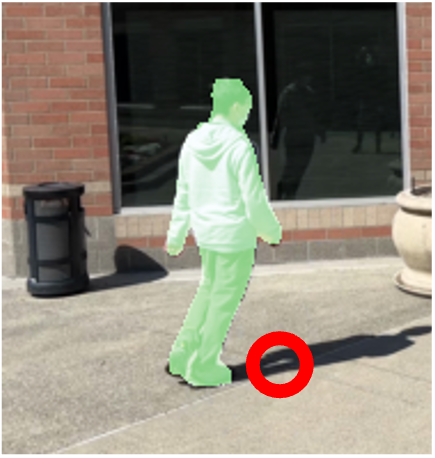} & 
\includegraphics[width=\width]{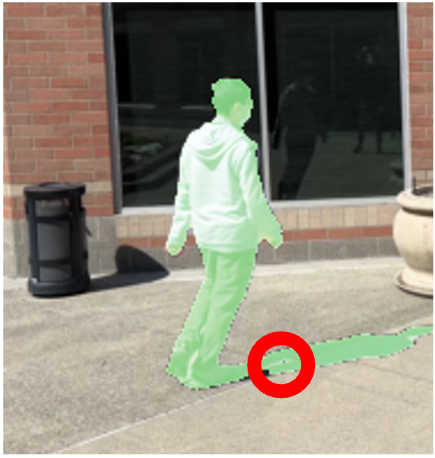} & 
\includegraphics[width=\width]{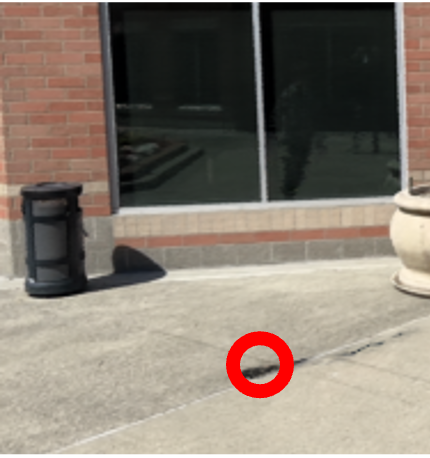} \\
{\footnotesize (a)} & {\footnotesize (b)} & {\footnotesize (c)}
\end{tabular}
\caption{
\textbf{Background Layer Training Signals.} We illustrate how the training signal to the background layer changes over time. It explains why the background captures some of the associated effects (in this example, shadows). We use the pixel circled in {\color{red}red} as an example. (a) At the beginning of training, the foreground alpha value (in light green) does not include the shadow. Therefore, $\alpha$ is small and at this pixel, $\hat{I}_t(x,y)\approx B_t(x,y)$. The reconstruction loss $\Lrecons$ encourages the background network $\bgnet$ to produce dark prediction at this location from this viewing angle. (b) As training progresses, $\alpha$ gets larger in the shadow region, and $\hat{I}_t(x,y)\approx C_t^i(x,y)$. This means that $\bgnet$ receives little to no supervision signals from this pixel. If it has modeled the shadow in some ways (in this case, a hole), it has little incentive to remove it, leaving the artifact in (c).
}	
\label{fig:bg_signals}
\end{figure}

When the pipeline is trained jointly as described above, it is sometimes observed that the background radiance field models some of the foreground contents like shadows (see Fig. \ref{fig:bg_signals}(c)). Compared to 2D images, 3D radiance fields are so much more capable that they can exploit distorted geometry constructs, such as holes and floaters, to capture some temporal effects, although the models are given no time information. For example, as the camera moves over time, there may be a correlation between whether a surface is covered by shadow and the direction the surface is viewed from.

We illustrate this problem in Fig. \ref{fig:bg_signals} and explain the cause at an intuitive level. The foreground branch is bootstrapped to produce alpha values that match the coarse mask inputs, which include only the object without the associated effects. In other words, $\alpha_t$ values are close to one at the object, but zero in the shadows (for simplicity, we consider one foreground layer in which the object casts a shadow, like in Fig. \ref{fig:bg_signals}). At a pixel $(x,y)$ covered by shadow, Eq. \ref{eq:compose} simply collapses to $\hat{I}_t(x,y)\approx B_t(x,y)$. The reconstruction loss will therefore encourage $B_t(x,y)$ to match the color of the shadow for a ray shot toward this location.

As training proceeds, $\fgnet$ will then gradually increase the predicted alpha values at the shadowed regions. If the shadow is hard and $\alpha$ gets close to one, Eq. \ref{eq:compose} evaluates to $\hat{I}_t(x,y)\approx C^i_t(x,y)$, and the reconstruction loss gives little to no constraint to the background color at the pixel. As a result, $\bgnet$ is unable to learn to remove the shadow color that it produces for the ray towards frame $I_t$ at $(x,y)$.

There are also cases where the shadow is soft and $\alpha$ is in between. In these cases, the problem remains ambiguous.

Therefore, we propose to obtain clean background reconstruction via an optional optimization step. In joint training, the foreground \textit{omnimatte} layers can capture most associated effects, including the parts with leaked content in the background layer. The alpha layers $\alpha_t$ can then be used to train a radiance field model from scratch, with no samples from the foreground region where alpha values are high. We show in the ablation study (see Fig. \ref{fig:bg_stages}) that this step produces cleaner background reconstruction for in-the-wild videos. As only the background is optimized, the process is fast and takes less than an hour to complete.

\section{Evaluation}
\label{sec:result}

We compare our quantitative and qualitative methods with Omnimatte and \dtnerf{} \cite{omnimatte,d2nerf}, which are state-of-the-art methods in 2D video matting and 3D video segmentation, respectively. In addition, we compare with Layered Neural Atlas (LNA) \cite{layeredNeuralAtlas}, which uses a deformable 2D background in contrast to Omnimatte's static image. 

\subsection{The Movies Dataset}

Quantitative evaluation of background segmentation requires a dataset with both input videos and ground-truth background imagery. Prior works primarily use datasets like CDW-2014 \cite{cdw2014}, which are limited to mostly static backgrounds and are not applicable to our settings. Recently, \dskubric{} is proposed in \dtnerf{}, which enables the evaluation of 3D background synthesis. However, these videos have relatively simple scenes and lighting. To facilitate the evaluation of video matting and background segmentation in challenging scenarios, we select six clips from three Blender movies in Blender Studio \cite{blenderStudio}. Compared to \dskubric{}, they feature more complicated scenes and lighting conditions, large nonrigid motion of the characters, and higher resolution. To ensure usability, we manually edit the camera trajectories so that there are sufficient camera motions and the actors have reasonable sizes. We render the clips with and without the actors to obtain input and ground truth for background reconstruction evaluation purposes. The camera poses are also exported.

\subsection{Experiment Setup}
We evaluate the performance of our proposed method on four datasets.

\begin{enumerate}
	\item \dsmovies{}: our novel challenging dataset.
	\item \dskubric{}: the dataset generated and used in \dtnerf, which consists of five scenes of moving objects from 3D Warehouse \cite{3dwarehouse} rendered with Kubric \cite{kubric}.
	\item \dsdavis{} \cite{Perazzi2016davis, pont2017davis}: short clips with moving foreground subjects, like humans, cars, and animals. This dataset is widely used to evaluate 2D-background matting methods \cite{omnimatte, layeredNeuralAtlas, deformableSprites}. 
	\item \dswild{}: in-the-wild sequences collected from the internet that are closer to casually captured videos, with natural and noisier camera motions, including translations and rotations, as well as objects at different distances from the camera. Naturally, these videos have backgrounds that are challenging for pure 2D methods.
\end{enumerate}
\dskubric{} and \dsmovies{} are synthetic datasets with clean background layer renderings available. Note that novel view synthesis is not the focus of our method, so we evaluate the background with input views. Both datasets have known camera poses and object masks which are used for training and evaluation.

\dsdavis{} and \dswild{} are real-world videos without clean background. Therefore, we only perform a qualitative evaluation to demonstrate the robustness of our method. For videos in \dswild{} we recover camera poses with COLMAP. For videos that COLMAP cannot process reliably, including \dsdavis{} videos, we use poses from RoDynRF \cite{liu2023robust}.

To obtain coarse object masks, we attempt to extract them with pre-trained object segmentation models from Detectron 2 \cite{wu2019detectron2}. In case it does not work, we use the Roto Brush tool in Adobe After Effects. Detailed procedures are described in the supplementary material. 
It takes about 10 minutes of manual effort to produce a 200-frame mask.

For all videos, we also estimate homographies with LoFTR \cite{sun2021loftr} and OpenCV to enable Omnimatte processing.

As mentioned in \dtnerf{} \cite{d2nerf}, the method is sensitive to hyperparameters. The authors released five sets of configurations for different videos. We experiment with every video using all provided configurations and report the best-performing ones.

\subsection{Implementation Details}

Our network is built upon the publicly available official implementation of Omnimatte \cite{omnimatte}, and TensoRF \cite{tensorf}. The videos in \dskubric{} have resolution $512\times 512$, and all methods run at the resolution $256\times 256$. For videos in other datasets with a higher resolution of $1920\times 1080$, we downsample them by a factor of 4.

We optimize the networks for up to 15,000 steps. 
The learning rate of $\fgnet$ is set to $0.001$ and is exponentially decayed after 10,000 steps. For $\bgnet$ we use the learning rate scheduling scheme of TensoRF. Training takes up to 6 hours on a single RTX3090 GPU. Detailed network architecture, hyper-parameters and timing data are presented in the supplementary. 
Our code and datasets will also be made publicly available.

\newcommand{\metricscell}[0]{\texttt{LPIPS}$\downarrow$ & \texttt{SSIM}$\uparrow$ & \texttt{PSNR}$\uparrow$}

\begin{table*}[t]
    \centering

    \mpage{0.98}{
    \small
    \resizebox{\linewidth}{!}{
    \begin{tabular}{c|ccc|ccc|ccc|ccc|ccc}
    	\toprule
   	    \multirow{2}{*}{\textbf{Kubrics}} & \multicolumn{3}{c|}{Car} & \multicolumn{3}{c|}{Cars} & \multicolumn{3}{c|}{Bag} & \multicolumn{3}{c|}{Chair} & \multicolumn{3}{c}{Pillow} \\
    	& \metricscell & \metricscell & \metricscell & \metricscell & \metricscell \\

\dtnerf{} & \second{0.135} & \second{0.854} & \second{34.10} & \second{0.105} & \second{0.859} & \second{34.77} & \second{0.131} & \second{0.880} & \second{33.98} & \second{0.090} & \second{0.916} & \second{33.29} & 0.105 & \second{0.926} & \second{38.80} \\
Omnimatte & 0.162 & 0.819 & 31.14 & 0.157 & 0.834 & 31.20 & 0.271 & 0.796 & 23.64 & 0.175 & 0.865 & 26.91 & 0.270 & 0.841 & 21.17 \\
LNA & - & - & -  & - & - & -  & 0.138 & 0.835 & 27.08 & 0.105 & 0.881 & 21.21 & \second{0.080} & 0.923 & 31.66 \\
Ours & \first{0.033} & \first{0.958} & \first{39.09} & \first{0.032} & \first{0.961} & \first{39.78} & \first{0.029} & \first{0.972} & \first{39.58} & \first{0.023} & \first{0.977} & \first{42.46} & \first{0.022} & \first{0.982} & \first{43.62} \\

   		\midrule
   		\multirow{2}{*}{\textbf{Movies}} & \multicolumn{3}{c|}{Donkey} & \multicolumn{3}{c|}{Dog} & \multicolumn{3}{c|}{Chicken} & \multicolumn{3}{c|}{Rooster} & \multicolumn{3}{c}{Dodge} \\
   		& \metricscell & \metricscell & \metricscell & \metricscell & \metricscell \\
   		
\dtnerf{} & - & - & -  & 0.370 & 0.694 & 22.73 & - & - & -  & 0.340 & 0.708 & 25.13 & 0.408 & 0.729 & 20.95 \\
Omnimatte & 0.315 & 0.653 & \second{19.11} & 0.279 & 0.706 & 21.74 & 0.312 & 0.704 & \second{20.95} & 0.220 & 0.741 & 23.14 & \second{0.067} & 0.879 & 23.88 \\
LNA & \second{0.104} & \second{0.849} & 18.79 & \second{0.154} & \second{0.828} & \second{26.08} & \second{0.190} & \second{0.818} & 19.22 & \second{0.131} & \second{0.804} & \second{26.46} & 0.068 & \second{0.937} & \second{24.94} \\
Ours & \first{0.005} & \first{0.990} & \first{38.24} & \first{0.030} & \first{0.976} & \first{31.44} & \first{0.021} & \first{0.978} & \first{32.86} & \first{0.024} & \first{0.969} & \first{27.65} & \first{0.006} & \first{0.991} & \first{39.11} \\

   		\bottomrule
    \end{tabular}
    }
    }

\caption{
    \textbf{Quantitative evaluations.} We present the background reconstruction comparison of our method and baselines on the \dskubric{} and \dsmovies{} datasets. Best results are in \first{bold} and second place are \second{underlined}. Results marked - are the ones the method failed to give good separations (visuals in supplementary).
    }
\label{tab:quantitative}
\end{table*}

\providelength\width
\setlength\width{1.8cm}
\begin{figure}
\footnotesize
\centering

\renewcommand{\tabcolsep}{1pt}
\begin{tabular}{ccccc}

& \dtnerf{} & Omnimatte & LNA & Ours \\

\input{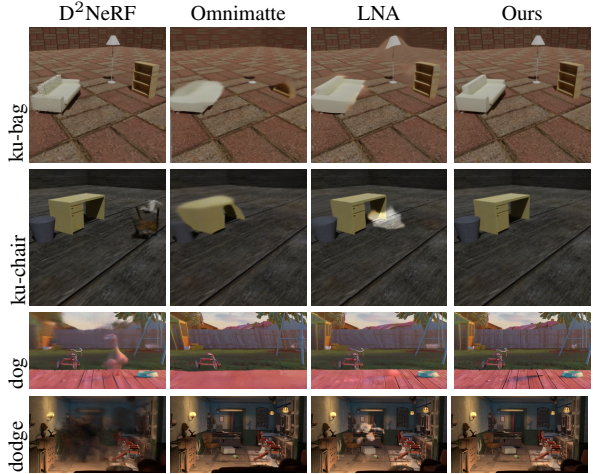}

\end{tabular}

\caption{
\textbf{Background Reconstruction.} We show examples of results presented in quantitative evaluations. For videos with parallax effects, 3D methods like \dtnerf{} and ours reconstruct less distorted background than Omnimatte and LNA.
}
\label{fig:bg_recons}
\end{figure}

\subsection{Quantitative Evaluation}

We quantitatively evaluate the background reconstruction quality of our method on two synthetic datasets. 
We report PSNR, SSIM and LPIPS for all videos in Table \ref{tab:quantitative}, and some visualizations in Fig. \ref{fig:bg_recons}.
For \dtnerf{}, we tried every provided pre-set configuration for every video in \dsmovies{}, and it only gave good results for the Dog, Rooster, and Dodge videos. Omnimatte and LNA with the 2D background layers struggles in both datasets. Our method can handle these videos well.

\providelength\width
\setlength\width{2.3cm}
\begin{figure*}
\centering

\renewcommand{\tabcolsep}{1pt}
\begin{tabular}{@{}cccccccccc@{}}

&&& \multicolumn{3}{c}{\footnotesize Background} && \multicolumn{3}{c}{\footnotesize Foreground} \\
\cmidrule{4-6}\cmidrule{8-10}
& {\footnotesize Input} && {\footnotesize \dtnerf{}} & {\footnotesize Omnimatte} & {\footnotesize Ours} && {\footnotesize \dtnerf{}} & {\footnotesize Omnimatte} & {\footnotesize Ours} \\

\input{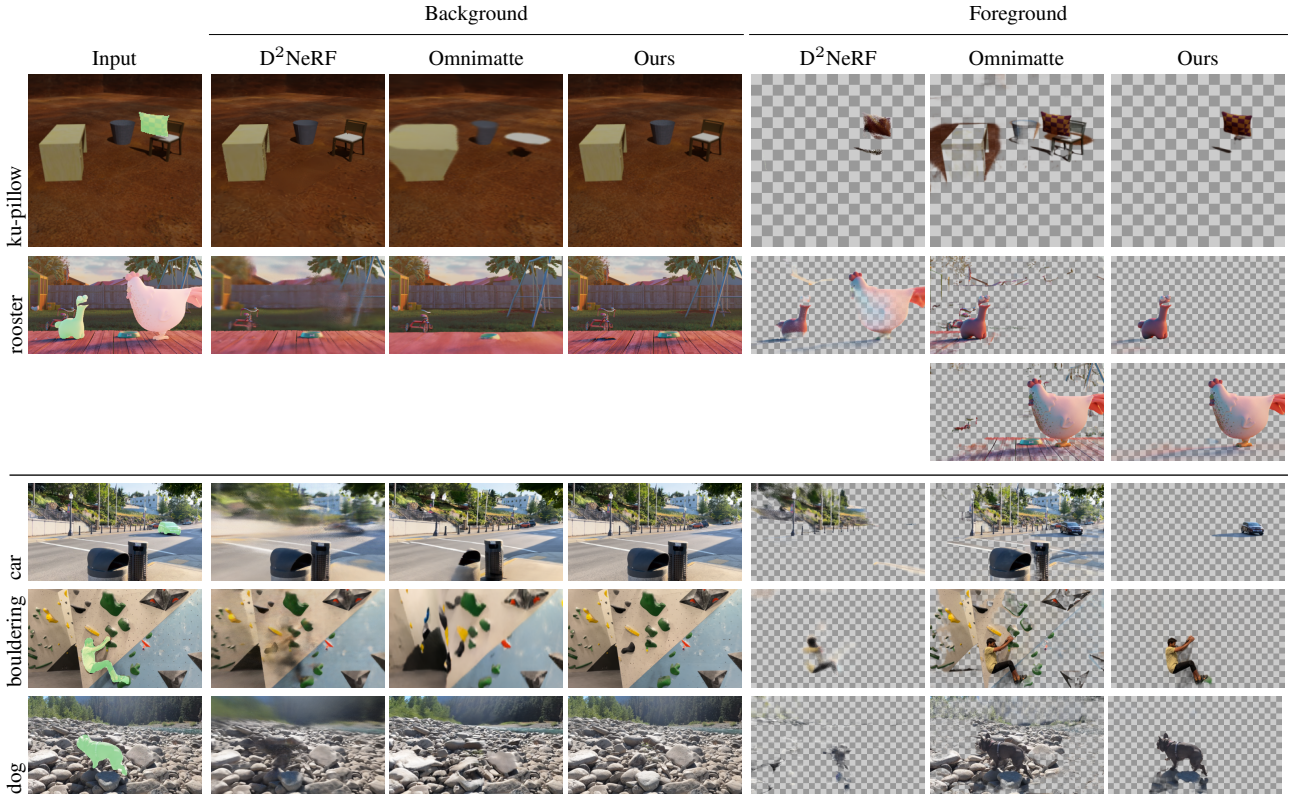}
\end{tabular}

\caption{ 
\textbf{Qualitative comparison.} We compare results of our and baseline methods on videos from each dataset. Readers are strongly encouraged to view videos files of more sequences available in the supplementary. The first two videos are synthetic from \dskubric{} and \dsmovies{}, followed by three \dswild{} videos. Omnimatte fails to handle objects in 3D and produces distorted background. \dtnerf{} works for videos with appropriate hyper-parameters, but does not generalize to new videos easily. Our method handles videos in many different settings. Due to space constraint we defer LNA results to the supplementary.
}	
\label{fig:qualitative}
\end{figure*}

\subsection{Qualitative Evaluation}

We present a qualitative comparison of the methods in Fig. \ref{fig:qualitative}. 
Due to space limitations, we present at least one video from every dataset but show a frame from every selected video in the figure. The original videos are available in supplementary and we highly recommend watching them. \dtnerf{} works well for the fine-tuned videos but not for new inputs without further hyper-parameter tuning. Omnimatte background has significant distortion around objects, and its foreground layer has to compensate for the limitation by capturing all residuals. Our method is versatile enough to perform well for a variety of videos with our 3D background model.

\subsection{Ablation Studies}

\providelength\width
\setlength\width{2.5cm}
\begin{figure}
\footnotesize
\centering

\renewcommand{\tabcolsep}{1pt}
\begin{tabular}{ccc}

Input & without $\Ldepth$ & with $\Ldepth$ \\

\input{images/monodepth_loss/source.tex} \\

Input & without $\Lflowrecons$ & with $\Lflowrecons$ \\

\input{images/flow_recons_legal/source.tex}

\end{tabular}

\caption{
\textbf{Loss Term Ablations.} Background of real-world videos without $\Ldepth$ and foreground without $\Lflowrecons$ can de degraded for real-world videos.
}
\label{fig:without_monodepth}
\end{figure}

\subsubsection{Loss Terms}

We present background reconstruction results without $\Ldepth$ in Fig. \ref{fig:without_monodepth}. For video sequences with rotational camera poses, the model struggles to extract 3D information from the input videos because of a lack of 3D clues. This loss is critical to extending our method to a broader range of videos. The effects of $\Lflowrecons$ are also demonstrated in Fig. \ref{fig:without_monodepth}. The auxiliary task improves foreground quality and reduces unrelated content.

\providelength\imgwidth
\setlength\imgwidth{2cm}

\begin{figure}
\centering
\small

\renewcommand{\tabcolsep}{1pt}
\begin{tabular}{ccc}

\input{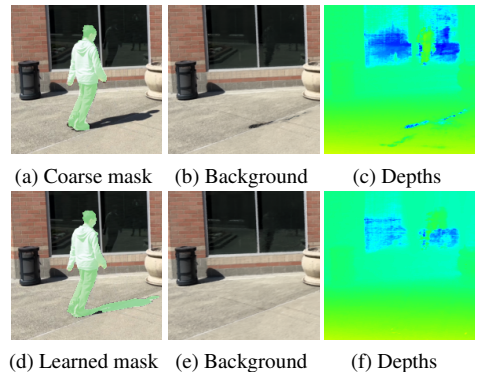}
\end{tabular}

\caption{
\textbf{Clean Background Retraining.} Background layers jointly trained can capture the shadows as a hole on the ground (a-c). After the joint training, the foreground \textit{omnimatte} provides a better mask that can be used to train a clean background (d-f). 
}	
\label{fig:bg_stages}
\end{figure}

\subsubsection{Clean Background Retraining}

We employ an additional step for real-world sequences to optimize a clean background from scratch. In Fig. \ref{fig:bg_stages}, we compare the background layer from the initial joint optimization and the final result. This is a simple yet robust way to obtain a better background.

\begin{figure}
\centering
\small

\renewcommand{\tabcolsep}{1pt}
\begin{tabular}{ccc}

\input{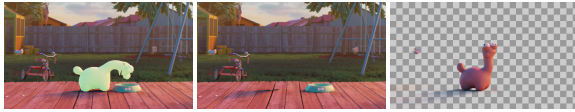} \\
{\footnotesize (a)} & {\footnotesize (b)} & {\footnotesize (c)}
\end{tabular}

\caption{
\textbf{Limitations.} (a), (b): Background can have baked-in shadows when the region is covered by shadows for most frames of the video. (c): The foreground layer captures irrelevant object motions (middle left) in the background. Best viewed in videos.
}	
\label{fig:failure}
\end{figure}

\subsection{Limitations}

We list some limitations that future works can explore.

\begin{enumerate}
\item If a background region is covered by shadows nearly all of the time, the background model cannot recover its color correctly. An example from a \dsmovies{} video is shown in Fig. \ref{fig:failure}. In theory, an \textit{omnimatte} layer has an alpha channel and can capture only the additive shadow that allows the background to have the original color. However, this problem is largely under-constrained in the current setting, making it ambiguous and leading the background to unsatisfying solutions.
\item The foreground layer captures irrelevant content. In real-world videos, unrelated motions often exist in the background, like swaying trees and moving cars. These effects cannot be modeled by the static radiance field and will be captured by the foreground layer regardless of their association with the object. Possible directions include i) using a dummy 2D layer to catch such content or ii) a deformable 3D background model with additional regularization to address the ambiguity as both background and foreground can model motion.
\item Foreground objects may have missing parts in the \textit{omnimatte} layers if they're occluded. Since our foreground network predicts pixel values for alpha composition, it does not always hallucinate the occluded parts.
\item The video resolution is limited. This is primarily due to the U-Net architecture of the foreground model inherited from Omnimatte. Higher resolutions can potentially be supported with the use of other lightweight image encoders.
\item The foreground layer may capture different content when the weights are randomly initialized differently. We include visual results in the supplementary materials.
\end{enumerate}

\section{Conclusion}
\label{sec:conclusion}

We propose a method to obtain \textit{omnimattes}, RGBA layers that include objects and their associated effects by combining 2D foreground layers and a 3D background model. Extensive experiments demonstrate that our approach is applicable to a wide variety of videos, expanding beyond the capabilities of previous methods.

{\small
\bibliographystyle{ieee_fullname}
\bibliography{egbib}
}

\newpage
\appendix

\renewcommand{\thefigure}{A\arabic{figure}}
\renewcommand{\thetable}{A\arabic{table}}
\setcounter{figure}{0}
\setcounter{table}{0}

\section{Additional Qualitative Results}

Video files of results presented in the main paper (all videos from the \dsmovies{}, \dskubric{}, \dswild{}, and \dsdavis{} datasets) are \updated{available on our project website as part of the supplementary material. We highly recommend watching them on: \url{https://omnimatte-rf.github.io}}

\begin{enumerate}
    \item For our method (OmnimatteRF), we include results (inputs with masks, foreground layers, background layer, background depth map) for every video.
    \item For \dtnerf{} \cite{d2nerf}, we use the best result among all configurations provided by the authors for every video. If none of the configurations successfully reconstruct non-empty static and dynamic layers, we drop the video files and only show a frame in Fig. \ref{fig:suppl_d2nerf}.
    \item For Omnimatte \cite{omnimatte} and Layered Neural Atlas (LNA) \cite{layeredNeuralAtlas}, \updated{we include videos from \dswild{}, \dsmovies{}, and \dskubric{}.} Results of \dsdavis{} can be found in  prior works.
\end{enumerate}

\section{Random Initialization}

\updated{As is also discussed in Omnimatte~\cite{omnimatte}, different random initializations can lead to varying results of the foreground layers. We show two examples in Fig. \ref{fig:suppl_seed}.

In all our experiments, the random seed is set to 3.
}

\providelength\width
\setlength\width{2.6cm}

\begin{figure}[h]
\centering
\small

\renewcommand{\tabcolsep}{1pt}
\begin{tabular}{ccc}

\includegraphics[width=\width]{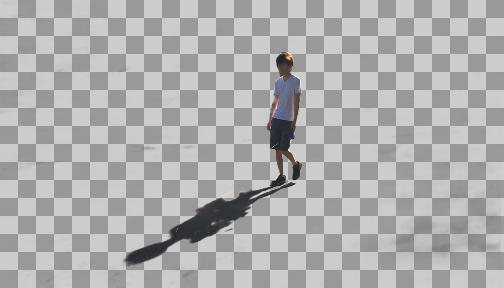} &
\includegraphics[width=\width]{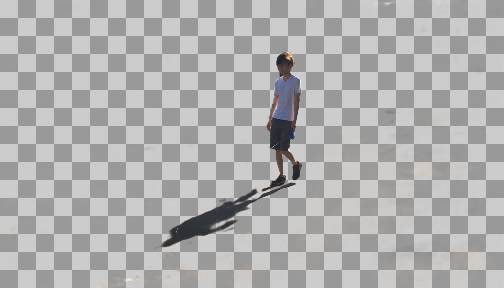} &
\includegraphics[width=\width]{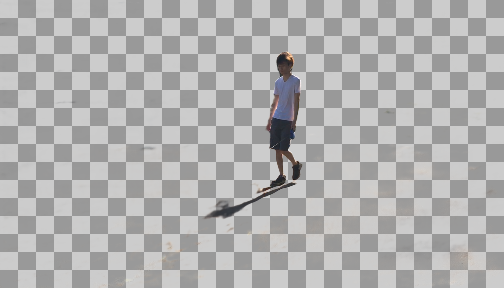} \\
{\footnotesize Seed = 5} & {\footnotesize Seed = 1} & {\footnotesize Seed = 0} \\
\cmidrule{1-3}
\includegraphics[width=\width]{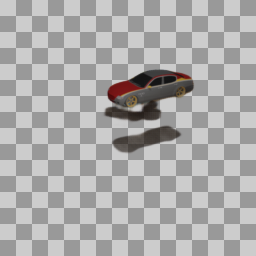} &
\includegraphics[width=\width]{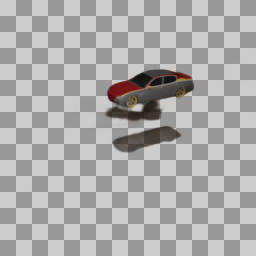} &
\includegraphics[width=\width]{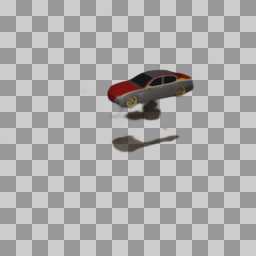} \\
{\footnotesize Seed = 3} & {\footnotesize Seed = 1} & {\footnotesize Seed = 0} \\
\includegraphics[width=\width]{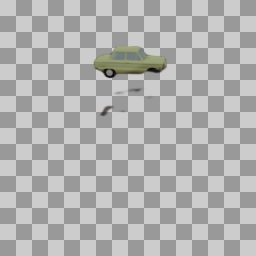} &
\includegraphics[width=\width]{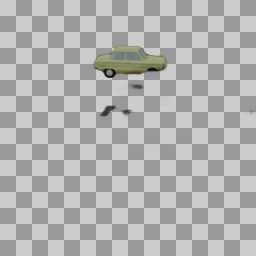} &
\includegraphics[width=\width]{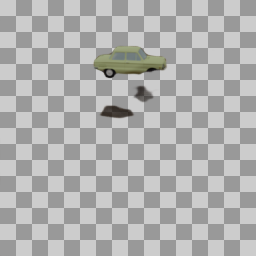} \\
{\footnotesize Seed = 3} & {\footnotesize Seed = 1} & {\footnotesize Seed = 0} \\
\end{tabular}
\caption{
\textbf{Effect of random initialization.} Top: for the \texttt{Wild/dogwalk} video, different seeds lead to different amount of hallucinated shadow of the person. Bottom: for the \texttt{Kubrics/cars} video, seeds influence how shadows are associated to the objects.
}

\label{fig:suppl_seed}
\end{figure}

\section{Additional Implementation Details}

\subsection{Mask Generation}

Our method and Omnimatte relies on coarse mask videos that outlines every object of interest. The synthetic \dskubric{} and \dsmovies{} videos have ground truth object masks and we use them directly. To obtan mask for an in-the-wild video, we use one of the two workflows:

\begin{enumerate}
    \item We first process the video a the pretrained Mask R-CNN model (\texttt{X101-FPN}) from Detectron 2 \cite{wu2019detectron2}. Then, we manually select a mask in every frame that best capture the object.
    \item We use the Roto Brush tool in Adobe After Effects to track the object. This method is useful when Mask R-CNN fails produce good masks for a video. \updated{In particular, we processed \texttt{Wild/dance} and \texttt{Wild/solo} manually.}
\end{enumerate}

It takes about 10 minutes of manual work to generate a mask sequence for a 200-frame video.

\subsection{Network Architecture}

Our foreground network is based on the U-Net architecture of Omnimatte, which is detailed in their supplementary \cite{omnimatte}. To adopt their network to OmnimatteRF, we replace the background noise input by the 2D feature map $E_t$. Each pixel in $E_t$ is the positional encoding of the 3D vector $(x,y,t)$ where $(x,y)$ is the pixel location and $t$ is the frame number. The positional encoding scheme is the same as proposed in NeRF \cite{mildenhall2020nerf}, with $L=10$ frequencies.

For background, we use the Vector-Matrix decomposition model in TensoRF \cite{tensorf} with the MLP feature decoder. Our initial grid has the same resolution $N_0=128$, and the final grid is limited to $N=640$. The vectors and matrices are upsampled at steps 2000, 3000, 4000, 5500.

\subsection{Hyper-parameters}

For all videos, we use a learning rate of 0.001 for the foreground network, which is exponentially decayed from the 10,000 step at a rate of $0.1\times$ per 10,000 steps. We find the decay crucial in preventing the foreground training from diverging. The mask bootstrapping loss $\Lmask$ has an initial weight of 50, which is first reduced to 5 when the loss value (before weighting) drops to below 0.02, and then turned off after the same number of steps. We document weights of other loss terms in Table \ref{tab:suppl_hyperparameters}.

Background network learning rate scheduling and $\Lbgreg$ weight are identical as the original TensoRF \cite{tensorf}.

In general, we use the same set of hyper-parameters for most videos, and only add additional terms when artifacts are observed.

\begin{table*}[h]
    \centering

    \mpage{0.98}{
    \footnotesize
    \begin{tabular}{cccccccc}
    \toprule
    Video & Steps & $\Lrecons$ & $\Lalphareg$ & $\Lalphawarp$ & $\Lflowrecons$ & $\Ldepth$ & $\Ldistort$ \\
    \midrule
    All & 15,000 & 1 & 0.01 (L1) / 0.005 (L0) & 0.01 & 1 & 0 & 0 \\
    Wild/bouldering & - & - & - & - & - & 0.1 & 0.01 \\
    \dsdavis{} & 10,000 & - & - & - & - & 1 & 0 \\
    \bottomrule
    \end{tabular}
    }

\caption{
    \textbf{Hyper-parameters.} We document the hyper-parameters (number of steps and weights of loss terms) in our experiments. The first row is the configuration shared by most videos. Remaining rows are videos with different configurations, and - means unchanged from the shared number.
    }
\label{tab:suppl_hyperparameters}
\end{table*}

\begin{table}[t]
    \centering

    \mpage{0.98}{
    \small
    \resizebox{\linewidth}{!}{
    \begin{tabular}{cccc}
    \toprule
    Method & Steps & Training (hours) & Rendering (s/image) \\
    \midrule
    Omnimatte & 12,000 & 2.7 & 2.5 \\
    \dtnerf{} & 100,000 & 4.5 & 4.8 \\
    LNA & 400,000 & 8.5 & 0.40 \\
    Ours & 15,000 & 3.8 & 3.5 \\
    \midrule
    Omnimatte & 12,000 & 1.2 & 0.95 \\
    \dtnerf{} & 100,000 & 3.2 & 2.2 \\
    LNA & 400,000 & 8.5 & 0.21 \\
    Ours & 15,000 & 2.5 & 1.2 \\
    \bottomrule
    \end{tabular}
    }
    }

\caption{
    \textbf{Running Time Measurement.} We measure and compare the time it takes to train OmnimatteRF and baseline methods. \textbf{Top}: \dsmovies{}, \dswild{} ($480\times 270$, \dsdavis{} has a similar resolution of $428\times 240$). \textbf{Bottom}: \dskubric{} ($256\times 256$).
    }
\label{tab:suppl_running_time}
\end{table}

\providelength\width
\setlength\width{2.6cm}
\begin{figure*}
\centering
\footnotesize

\renewcommand{\tabcolsep}{1pt}
\begin{tabular}{cccccccc}

FG & BG && FG & BG && FG & BG \\

\input{images/suppl_failure/source_d2nerf}

\end{tabular}

\caption{
\textbf{\dtnerf{} results for failed scenes.}
}
\label{fig:suppl_d2nerf}
\end{figure*}













\subsection{Running Time Measurement}

We measure and report the time it takes to train OmnimatteRF and baseline methods in Table \ref{tab:suppl_running_time}. All measurements are conducted on a workstation with an eight-core AMD R7-2700X CPU and a single NVIDIA RTX3090 GPU.

Our method takes longer to train than Omnimatte due to the addition of the 3D background radiance field.

\updated{Optimizing the background model only, as in the clean background retraining process, takes about 30 minutes per video.}

\end{document}